\pdfoutput=1

\documentclass[11pt]{article}

\usepackage[]{eacl}

\usepackage{times}
\usepackage{latexsym}
 \usepackage{enumitem}

\usepackage{xcolor}
\usepackage{microtype}
\usepackage{graphicx}
\usepackage{float} 
\usepackage{subfigure}
\usepackage{amsmath}
\usepackage{amssymb}

\usepackage{booktabs}
\usepackage{courier}
\setlist[itemize]{itemsep=1pt}

\usepackage[T1]{fontenc}

\usepackage[utf8]{inputenc}

\usepackage{microtype}

\usepackage{inconsolata}

\title{Advancing Precise Outline-Conditioned Text Generation with Task Duality and Explicit Outline Control}

\author{Yunzhe Li \\
University of Illinois, Urbana-Champaign\\
\texttt{yunzhel2@illinois.edu} \\
\And
Qian Chen \\
Speech Lab, Alibaba Group \\
\texttt{tanqing.cq@alibaba-inc.com} \\
\AND
Weixiang Yan \\
University of California, Santa Barbara\\ 
\texttt{weixiangyan@ucsb.edu} \\
\And
Wen Wang \\
Speech Lab, Alibaba Group \\
\texttt{w.wang@alibaba-inc.com} \\
\AND
Qinglin Zhang \\
Speech Lab, Alibaba Group \\
\texttt{qinglin.zql@alibaba-inc.com} \\
\And
Hari Sundaram \\
University of Illinois, Urbana-Champaign\\
\texttt{hs1@illinois.edu}\\}

\begin{document}
\maketitle
\vspace{5cm}
\begin{abstract}
Existing works on outline-conditioned text generation typically aim to generate text using provided outlines as rough sketches, such as keywords and phrases. However, these approaches make it challenging to control the quality of text generation and assess consistency between outlines and generated texts due to lack of clarity and rationality of the rough outlines. In this paper, we introduce a novel text generation task called \textbf{\textit{Precise Outline-conditioned Generation}}, which requires generating stories based on \textit{specific}, \textit{sentence-level} outlines. To facilitate research on this task, we construct two new datasets, \textbf{WPOG} and \textbf{CDM}. We provide strong baselines based on fine-tuning models such as BART and GPT-2, and evaluating zero-shot performance of models such as ChatGPT and Vicuna. Furthermore, we identify an issue of \textbf{imbalanced utilization of the outline information} in the precise outline-conditioned generation, which is ubiquitously observed across fine-tuned models and zero-shot inference models. To address this issue, we propose an \textbf{explicit outline utilization control approach} and a novel framework that \textbf{leverages the task duality between summarization and generation}. Experimental results show that the proposed approaches effectively alleviate the issue of imbalanced outline utilization and enhance the quality of precise outline-conditioned text generation for both fine-tuning and zero-shot settings.\footnote{Implementation is released at https://github.com/yun\\zhel2/precise-outline-gen.}

\end{abstract}

\begin{figure}[t]
    \centering
    \setlength{\belowcaptionskip}{-0.5cm} 
    \includegraphics[width=0.45\textwidth]{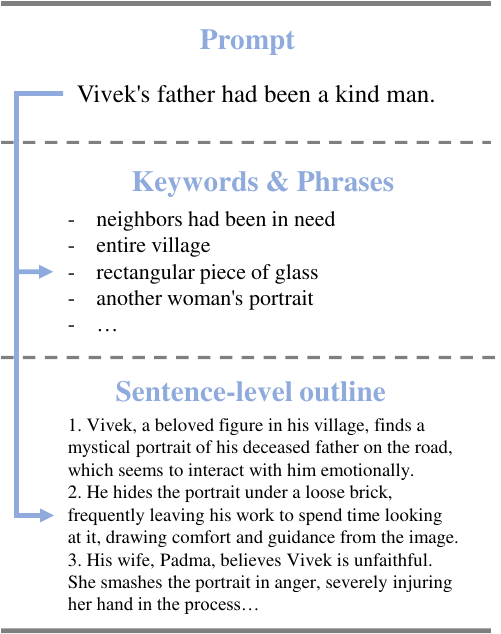}
    \caption{The comparison between different forms of outlines. Given the prompt (e.g., title), the outline could be formulated as a) \textbf{rough} outlines: a list of keywords or phrases, and b) \textbf{precise} outlines: salient sentence-level statements.}
    \label{fig:question}
\end{figure}

\section{Introduction}
\label{sec:introduction}

Outline-conditioned text generation is a challenging but important task, which provides structured conditions (outlines) that aim to enhance content quality, customization, and generation efficiency for a variety of downstream text generation applications, e.g., story-telling and long-form question-answering. It requires the model to not only generate fluent and coherent text but also ensure that the generated text aligns with the content, structure, and properties (e.g., style) specified in the outline.  Existing works~\citep{fan2018hierarchical, rashkin2020plotmachines, fang2021outline} mainly regard a set of keywords or phrases as the ``rough outline''. The rough outline is usually extracted via heuristic methods~\citep{rose2010automatic,campos2018yake} and is overly abstractive, making it difficult to guarantee either quality or rationality of the rough outline, which in turn hampers quality of the generated text conditioned on the outline. Moreover, the ambiguity and poor readability of the rough outline also complicate the evaluation of consistency between the outlines and generated text (e.g., simply asking the evaluator ``Which text is better at utilizing the keywords?'' as in~\citep{rashkin2020plotmachines} could be quite challenging to answer).

To address these limitations of text generation conditioned on rough outlines, we introduce a novel task called \textbf{\textit{Precise Outline-conditioned Generation}}. As shown in Figure~\ref{fig:question}, this task takes the specific, sentence-level outline rather than keywords and phrases, which could improve the readability, clarity, and rationality of outlines and hence the quality of generated text and could better facilitate assessing consistency between the outline and generated text. Compared to text generation conditioned on rough outlines, our proposed task poses greater challenges in three aspects: 
a) \textbf{Controllability:} Precise outlines inherently encode properties (e.g., style and attributes) that vary drastically with scenarios (e.g., stories, news reports). Models need to grasp the properties and treat them as pre-given constraints to control text generation. b) \textbf{Faithfulness:} The generated text needs to be faithful to the outline in terms of not only content but also relations among content in the outline. c) \textbf{Structure:} The generated text needs to maintain a reasonable structure so that the key points in the generated text are well organized. 


To facilitate research on this new task, we provide two datasets \textbf{WPOG} and \textbf{CDM} from storytelling and news report scenarios, respectively. We study how competitive models address these challenges by fine-tuning widely used pre-trained language models (LMs) such as BART~\citep{Lewis2020BARTDS} and GPT-2~\citep{radford2019language} for inference and also zero-shot inference with top-performing large language models (LLMs) such as ChatGPT~\citep{ouyang2022training} and Vicuna~\citep{vicuna2023}\footnote{For LLMs with strong zero-shot inference ability~\cite{kojima2023large}, in this work, we only focus on in-context learning without further instruction tuning or other fine-tuning.}. The two datasets exhibit two drastically different styles, so we could evaluate how well models understand the inherent style and control generation conditioned on the style, i.e., handling the \textbf{controllability} challenge. We observe a common issue of imbalanced utilization of the outline across fine-tuning and zero-shot settings, which affects the structure of the generated text.
Specifically, the outline information is mostly used at the beginning of text generation and does not effectively guide the rest of the generation process. This issue indicates that the models struggle with the challenges of maintaining \textbf{faithfulness} and \textbf{structure}. We hypothesize that the imbalanced outline utilization issue may be caused by a combination of exposure bias due to teacher forcing (the model relies too heavily on previously predicted words) and decoding strategies (the model assigns the highest probabilities to the correct outline information), leading to \textbf{early depletion} of the outline. In order to accurately assess the faithfulness and structure quality of the generated text, we propose three metrics: \textbf{distribution variation}, \textbf{peak-value distance}, and \textbf{consistency degree}.

We propose two general approaches to mitigate this imbalance issue. The first method explicitly controls the usage of the given outline throughout the generation process. The second method exploits the task duality between precise outline-conditioned generation and summarization and implements a dual learning strategy.
Experimental results on the two datasets with automatic and human evaluations show that both approaches alleviate the imbalance issue and improve generation quality.



Our contributions can be summarized as follows: (1) We introduce a novel 
\textbf{Precise Outline-conditioned Generation} task. (2) We provide two datasets \textbf{WPOG} and \textbf{CDM} to support research on this task. (3) We identify the \textbf{imbalanced outline utilization issue} and propose three evaluation metrics to gauge outline utilization. (4) To alleviate this imbalanced outline utilization issue, we propose two \textit{general} approaches, \textbf{explicit outline utilization control} and \textbf{unified dual-task learning}. 
(5) Experimental results demonstrate that our approaches effectively mitigate the imbalanced outline utilization issue and improve the generation quality \textbf{for both fine-tuning and zero-shot settings}.

\section{Related Work}
\paragraph{Controllable Text Generation}
Controllable text generation is a core issue in text generation, which is also a requirement in virtually every context involving text generation \citep{Zhang2022ASO}. Previous works proposed a lot of related tasks to evaluate the controllability from coarse to fine-grained. Attribute-based generation requires the attributes-specific sentence with given topics \citep{Dathathri2019PlugAP, Wang2019TopicGuidedVA}, keywords \citep{Zhang2020POINTERCP, Carlsson2022FineGrainedCT} or sentiments \citep{He2020APF,Zhang2022DisCupDC}. However, the sentence-level target has less requirement for coherence or structure. For longer text generation, storytelling, and data-to-text aim to generate paragraphs or passages based on given topics and endings \citep{fan2018hierarchical, GoldfarbTarrant2020ContentPF, Tambwekar2018ControllableNS}, or tables and graph \citep{Su2021PlanthenGenerateCD, Ribeiro2020InvestigatingPL}. The closest task to ours is planning-based generation, which we call rough outline-conditioned generation. The main difference is that they primarily use keywords or phrases as conditions, which limits the requirements for generating results and assessing accuracy due to the vagueness of the conditions themselves. On the other hand, the rough outline is usually extracted by heuristic methods (e.g., RAKE~\citep{rose2010automatic} and YAKE~\cite{campos2018yake}) to select keywords from ground truth text by word frequency and graph metrics, which also limits the quality of data. 

\paragraph{Outline-conditioned Generation}
There are a variety of works that take outline-conditioned generation as the critical step to long text generation. \citet{Xu2018ASM, Yao2019PlanAndWriteTB, Shao2019LongAD} adopt the Bi-LSTM as the backbone with reinforcement learning and variational inference techniques to optimize with keywords/phrases skeleton, respectively. \citet{Fan2019StrategiesFS} modeling the semantic role labels (SRL) as outlines by the self-attention mechanism and pointer mechanism \citep{Vinyals2015PointerN}. 
\citet{Tan2020ProgressiveGO, Chen2021GraphPlanSG, Wang2022LanguageMV} use pretrained language models to tackle with outlines in the form of keywords, event-graph, and latent variables, respectively. 
\citet{sun2022summarize} propose a method to learn outline generation by reconstructing the summary, followed by generating segment-level text and concatenating them to obtain the full text.
For zero-shot inference with LLMs, most existing works focus on improving the interaction or fluency with self-generated outlines. \citet{Yang2022Re3GL, zhou2023recurrentgpt} apply the recursive prompting and revision to enhance GPT-3 and ChatGPT over long-form storytelling, respectively. \citet{Yang2023DOCIL} make a detailed outliner and a controller based on FUDGE~\citep{Yang2021FUDGECT} to improve long-range plot coherence. In this paper, we consider both fine-tuning and zero-shot settings and propose unified strategies that could work for both scenarios. 


\section{Methods}
\label{sec:method}
\subsection{Precise Outline-conditioned Generation}
\label{sec:generation}
\noindent \textbf{Baseline Approach} Our proposed precise outline-conditioned generation task takes specific sentence-level outlines and prompts as input, and requires generating long texts that are fluent, coherent, and match the input information.  Specifically,  the writing prompt (e.g., title or opening) is denoted by $\mathbf{x}=\{x_1,x_2,...,x_l\}$, where $x_i$ denotes the $i$-th token in the writing prompt. The outline $\mathbf{o}$ is a set of sentences $\mathbf{o}=\{o_1,o_2,...,o_m\}$, where $o_j$ is the $j$-th sentence and $o^t$ is the $t$-th token in $\mathbf{o}$. Given $\mathbf{(x, o)}$ as input, a model is expected to output $\mathbf{y}=\{y_1,y_2,...,y_n\}$, where $y_k$ denotes the $k$-th sentence and $y^t$ is the $t$-th token in $\mathbf{y}$.



We study two distinct approaches for implementing $\mathbf{(x,o)} \rightarrow \mathbf{y}$: \textbf{fine-tuning pre-trained LMs} and \textbf{zero-shot inference by LLMs}. For fine-tuning pre-trained encoder-decoder and decoder-only pre-trained LMs and then inference, the output is generated by the model $\theta$ as follows:
\begin{equation}
P\left(\mathbf{y} \mid \mathbf{o,x};\theta\right) = \prod_{t=1}^{n} P\left(y^t \mid y^{<t}, \mathbf{o,x};\theta\right) 
\end{equation}
An encoder-decoder model (e.g., BART) encodes the outline $\mathbf{o}$ by its encoder and treats the prompt $\mathbf{x}$ as the prefix for its decoder; whereas, a decoder-only model (e.g., GPT-2) treats both the outline and the prompt as the prefix for its decoder. 

For LLM zero-shot inference, we adopt \textit{prompt learning} to infer with the concatenation of the outline and the prompt. Details of prompt designing are in Appendix~\ref{subsec:allin}.

\noindent \textbf{Imbalanced outline utilization} We identify a common issue of imbalanced outline utilization in fine-tuning both encoder-decoder and decoder-only models as well as zero-shot inference. We conduct case studies in Table ~\ref{tab:example} (Appendix~\ref{sec:example}) and a similarity visualization in Figure~\ref{fig:dis} (Appendix~\ref{sec:analysis_appendix}) to further illustrate the problem. Specifically, the models tend to repeat all the outlines at the head of the output and then continue writing while only treating the output prefix as context. This issue indicates weaknesses of the models for handling \textbf{Faithfulness} and \textbf{Structure} challenges (Section~\ref{sec:introduction}). Lack of constant guidance from the outline to regulate the remaining writing may cause deviations from the outline and unfaithful content, resulting in poor faithfulness. Also, with the skewed distribution of repeating all the outlines at its head, the generated text could be ill-structured. We propose three metrics to evaluate outline utilization in Section~\ref{sec:metric}. To address this issue, we propose an explicit outline utilization control method and a unified dual learning strategy. 



\subsection{Explicit Outline Utilization Control}
\begin{figure}[ht]
    \centering
    \setlength{\belowcaptionskip}{-0.5cm} 
    \includegraphics[width=0.45\textwidth]{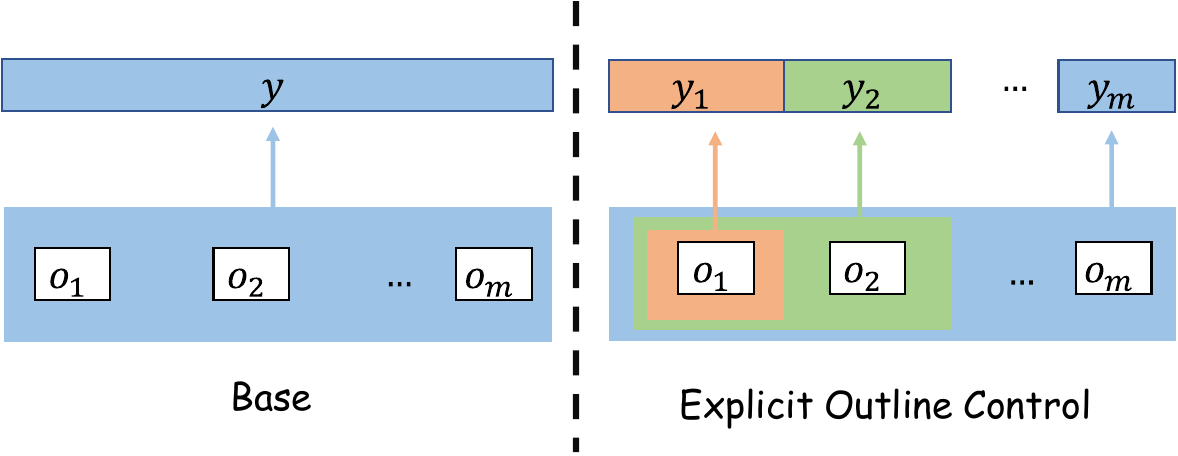}
    \caption{The proposed \textbf{Explicit Outline Utilization Control method} versus the baseline approach. We omit the prompt $\mathbf{x}$ for simplicity.}
    \label{fig:mode}
\end{figure}

As Figure~\ref{fig:mode} illustrates, we propose an approach that explicitly controls the utilization of the outline during generation by segmenting the outline into points (taking one outline sentence as one point in this work), generating text corresponding to each point, and aggregating and refining the outputs to obtain the final result. We denote this new method as \textbf{explicit outline utilization control (OC)}. OC can be used in both fine-tuning and zero-shot inference settings.


\paragraph{Fine-tuning}
We apply OC during both fine-tuning and inference. For each story $\mathbf{y}$ and its corresponding outline $\mathbf{o}=\{o_1,o_2,...,o_m\}$ in the training set, we partition $\mathbf{y}$ into $m$ snippets $\mathbf{s}=\{s_1,s_2,...,s_m\}$ and establish a mapping between each point $o_i$ in $\mathbf{o}$ and each snippet in $\mathbf{s}$ in order since the outline contains sequential information.
\textbf{In the fine-tuning stage}, we only calculate the loss from the \textbf{current} paired outline point and story snippet, while masking the other pairs. Hence, story partitioning may significantly impact the effectiveness of OC. We investigate two methods: \textbf{1) Average partition}, which divides stories by a constant number of sentences, and \textbf{2) Greedy Search}, which partitions stories by minimizing the distances between outline points and story snippets as well as the variance of the lengths of snippets. The objective function is as follows:
\begin{equation}
    \label{eq:gs}
    \mathcal{L}_{gs}= \alpha \sum_{i=1}^m \sum_{j \in s_i} dist(o_i,y_j) + \beta \sum_{i=1}^m (|s_i|-\mu_s)^2
\end{equation}
where $dist(\cdot)$ is the distance function, $|s_i|$ and $\mu_s$ denote the number of sentences in the story snippet $s_i$ and the mean of $|s_i|_{i=1}^{m}$. $\alpha$ and $\beta$ are weighting parameters. The second term of Eq.~\ref{eq:gs} is added to regularize the length distribution of story snippets.
\textbf{In the inference stage}, we first use the prompt and $o_1$ to generate the first part. Then we iteratively generate each part based on the next outline sentence and the generated context. Specifically, for encoder-decoder models, we adjust the outline for the encoder input and the decoder prefix with context; whereas for decoder-only models, we concatenate each outline sentence and already generated parts as input. All generated parts are aggregated as the final result.
\paragraph{Zero-shot inference} Similarly, we first use the prompt and $o_1$ with constraints including the expected length of the final story and the type (e.g., news article, story) to generate the first part. Then we iteratively generate each part based on the next outline sentence and the generated context. Finally, we aggregate all multi-round outputs. Since LLMs forget the previous settings in the multi-round interactions, we refine the aggregated output by reminding LLMs of the constraints and generation goal to obtain the final story. Details of prompts and examples are in Appendix~\ref{sec:separate}.

\subsection{Unified Dual-task Learning} 
Summarization can be regarded as the inverse process of precise outline-conditioned generation. We propose a unified approach leveraging this task duality (denoted \textbf{Dual}) for both fine-tuning and zero-shot inference settings for precise outline-conditioned generation, to alleviate the imbalanced outline utilization issue and in turn improve generation quality. 
\paragraph{Fine-tuning} We introduce a dual task in the fine-tuning stage for the generation model $\theta_{g}$, which summarizes the generated story with a model $\theta_{s}$. We consider both generation by the summarization model, $P(\mathbf{o} \mid \mathbf{y}; \theta_{s})$, and generation by the outline-conditioned generation model, $P(\mathbf{y} \mid \mathbf{o}; \theta_{g})$, generally satisfy Eq.\ref{eq:pc} based on Bayes’ rule. We omit the prompt $\mathbf{x}$ for simplicity.
\begin{equation}
\label{eq:pc}
P(\mathbf{o}) P(\mathbf{y} \mid \mathbf{o})= P(\mathbf{o}, \mathbf{y})= P(\mathbf{y}) P(\mathbf{o} \mid \mathbf{y})
\end{equation}
Inspired by~\citet{wei2019code}, we establish the bridge between the two tasks by adding a dual Lagrange regularizer term as Eq.\ref{eq:dual_loss} to the standard fine-tuning loss. It aims to utilize the shared knowledge between the dual tasks to strengthen their connection by minimizing the difference between the model output and their approximate probability constraints.
\begin{equation}
\label{eq:dual_loss}
\begin{aligned}
\mathcal{L}_{\text{dual}} = \big[&\log \hat{P}(\mathbf{o})  + \log P(\mathbf{y} \mid \mathbf{o} ; \theta_{g}) \\
&-\log \hat{P}(\mathbf{y}) -\log P(\mathbf{o} \mid \mathbf{y} ; \theta_{s})\big]^2
\end{aligned}
\end{equation}
where $\hat{P}(\mathbf{o}),\hat{P}(\mathbf{y})$ denote the marginal distributions of outlines and stories, respectively.

\paragraph{Zero-shot inference} We propose to leverage the task duality to refine LLM zero-shot inference via in-context learning. Specifically, an LLM first generates a draft story based on the given outline. Then the model is required to summarize the generated draft. Based on the property of duality, the given outline and the generated summary should cover equivalent information. Therefore, the model is asked to compare the outline and the summary to reflect and further refine the generated text based on the discovered discrepancy. Importantly, we also investigate combining Dual with OC (denoted by \textbf{OC+Dual}) by conducting the summarization step in Dual on the story parts generated under OC. The details of the prompt and examples of OC+Dual are shown in Appendix~\ref{subsec:dual} and~\ref{subsec:combo}.

\section{Dataset Creation}
To facilitate the study on the precise outline-conditioned generation task, we introduce two datasets, namely, WritingPrompt for Outline-conditioned Generation (\textbf{WPOG}) and CNN/DailyMail (\textbf{CDM})~\citep{Hermann2015TeachingMT}, from two distinct storytelling and news report scenarios, respectively. 


\noindent \textbf{WPOG} is constructed based on the English WritingPrompt (WP) dataset~\cite{fan2018hierarchical}, which includes human-written stories paired with writing prompts from an online forum. Since the original WP does not provide outlines, we randomly select 5\% of the samples and use GPT-4\footnote{https://openai.com/research/gpt-4} to generate the outlines for original stories. Moreover, we conduct the human evaluation to verify the reliability of generated outlines (the details and results are shown in Appendix~\ref{sec:data_valid}). The WPOG dataset is then built by pairing the generated outlines and the original writing prompts as input and the human-written stories as target references.
To ensure fair comparisons between fine-tuning and zero-shot inference settings, we consider the limitations on the number of tokens output by BART and GPT-2 and exclude samples longer than 1024 tokens.  We partition WPOG into 80\% for training, 10\% for validation, and 10\% for testing.


\noindent \textbf{CDM} is an English dataset containing unique news stories (written by journalists with CNN and the Daily Mail) and the corresponding human-written highlights. For our task, the highlights are taken as outlines to represent the key plots of the news reports. We keep the original partition of training and validation sets as~\citet{nallapati2016abstractive}, while sampling 10\% from the original test set as our test set for reasonable turnaround time for zero-shot evaluations of LLMs. We follow the previous work~\citep{fang2021outline} to obtain prompts and combine the prompt and outlines as input. 

\begin{table}[ht]
    \centering
    \renewcommand{\arraystretch}{0.90}
    \caption{Statistics of the CDM and WPOG datasets. Avg. length denotes the number of tokens from the BART tokenizer~\cite{Lewis2020BARTDS}.}
    \label{tab:dataset}
    \setlength{\tabcolsep}{4pt} 
    \begin{tabular}{lcc}
        \toprule  
        \textbf{Dataset} & \textbf{CDM} & \textbf{WPOG} \\
        \midrule
        \# Train & 287,113 & 6,982 \\
        \# Valid & 13,378 & 866  \\
        \# Test  & 1,149  & 866  \\
        Avg. length of story & 780.62 & 542.57 \\
        Avg. length of outline & 56.20 & 165.12 \\
        \bottomrule
    \end{tabular}
\end{table}



\section{Experiments}



The details of dataset preprocessing, dataset statistics, experimental setup, and implementations are presented in Appendix~\ref{sec:aes}.

\subsection{Baselines}
We select two baseline models for each scenario. For the fine-tuning scenario, we choose the typical encoder-decoder model \textbf{BART}~\citep{Lewis2020BARTDS} and decoder-only model \textbf{GPT-2}~\citep{radford2019language}. For zero-shot inference models, we select the state-of-the-art open-sourced model \textbf{Vicuna-v1.5}~\citep{vicuna2023} and the close-source model \textbf{ChatGPT}~\citep{ouyang2022training}.\footnote{It is noted that the current leading story generation and summarization models are not suitable for comparison, as their objectives and input requirements differ from this task. More details about baseline selection is in Appendix.~\ref{sec:aes}}

\subsection{Evaluation Metrics}
\label{sec:metric}
For evaluations, we adopt three metrics that are commonly used for text generation tasks. \textbf{Rouge (R-n)} evaluates the n-gram recall between generated texts and human-written texts, where n can be 1, 2, or L (longest common subsequence)~\citep{lin2003automatic}. \textbf{BERTScore} computes the F1-score between generated texts and human-written texts, using BERT~\citep{Devlin2019BERTPO} as a similarity measure~\citep{Zhang2019BERTScoreET}. \textbf{Distinct-4 (D-4)} measures the generation diversity (informativeness), based on the ratio of distinct 4-grams to all the generated 4-grams~\citep{Li2015ADO}.

We also propose three new metrics that measure how imbalanced the outline utilization is for precise outline-conditioned generation. 
\textbf{Distribution Variation (DV)} measures the difference between the distribution ($D_j$) of similarity between every outline sentence $o_j$ and generated text $\mathbf{y}$, which is calculated as Eq. \ref{eq:dv}, where $d_{KL}$ denotes KL-divergence between two distributions. 
\begin{equation}
      DV =\frac{1}{|\textbf{o}|\cdot(|\textbf{o}|-1)}\sum_{a,b\in \textbf{o},a\neq b}d_{KL}(D_a,D_b) 
      \label{eq:dv}
\end{equation}
\noindent \textbf{Peak-value Distance (PD)} indicates the distance of the most matched sentences ($P_j \in {1, 2, \dots, k}$) among the generated text $\mathbf{y}$ for every outline sentence $o_j$, respectively. It is defined as Eq.\ref{eq:pd}. 
    \begin{equation}
        PD =\frac{1}{|\textbf{o}|\cdot(|\textbf{o}|-1)}\sum_{a,b\in \textbf{o},a\neq b}|P_a-P_b| 
        \label{eq:pd}
    \end{equation}
\noindent \textbf{Consistency Degree (CD)} is proposed to assess outline utilization, motivated by the duality of summarization and precise outline-conditioned generation. Specifically, we use ChatGPT to summarize the generated text and contrast the summary with the original outline by computing Rouge-L of it against the outline as Consistency Degree (CD). The prompt used for CD is shown in Appendix~\ref{subsec:cd}. 

DV and PD could be considered as intrinsic metrics for outline utilization while CD is extrinsic. All of them are assessed from both faithfulness and structure perspectives (Section~\ref{sec:introduction}); and for each of them, a higher value suggests a more even outline utilization.


\subsection{Automatic Evaluation}
\begin{table*}[ht]
    \setlength{\belowcaptionskip}{-0.2cm} 
    \renewcommand{\arraystretch}{0.65}
    \centering
    \scalebox{0.85}{
    \begin{tabular}{lcccccccc}
        \toprule
        Method & R-1 & R-2 & R-L & D-4 & BERTScore & DV & PD & CD \\
        \midrule
        \multicolumn{9}{c}{\textbf{CDM}} \\
        \midrule
        BART-base & 38.79 & 10.86 & 16.26 & 79.47 & 84.62 & 2.21 & 1.09 & 14.74 \\
        w/ OC & \textbf{40.92} & 11.82 & \textbf{17.48} & 87.94 & 86.76 & \textbf{2.94} & \textbf{4.06} & \textbf{21.62} \\
        w/ Dual & 39.66 & 11.21 & 16.67 & 82.73 & 85.17 & 2.43 & 2.37 & 15.20 \\
        w/ OC + Dual & 40.47 & \textbf{12.04} & 17.30 & \textbf{89.26} & \textbf{87.21} & 2.66 & 2.93 & 17.39 \\
        \midrule
        ChatGPT-3.5-turbo & 41.17 & 12.17 & 17.62 & 97.46 & 84.97 & 2.63 & 3.44 & 18.96 \\
        w/ OC & 41.69 & 12.78 & 18.67 & \textbf{99.47} & 85.48 & 2.75 & 4.46 & \textbf{23.88} \\
        w/ Dual & 42.46 & \textbf{13.10} & \textbf{19.06} & 98.82 & 84.30 & 2.49 & 3.62 & 18.54 \\
        w/ OC + Dual & \textbf{42.90} & 12.82 & 18.89 & 97.83 & \textbf{87.29} & \textbf{2.85} & \textbf{4.76} & 19.40 \\
        \midrule
        Ground Truth & - & - & - & 89.24 & - & 3.16 & 8.67 & 29.62 \\
        \midrule[0.5pt]
        \multicolumn{9}{c}{\textbf{WPOG}} \\
        \midrule
        BART-base & 39.20 & 14.19 & 22.08 & 92.89 & 83.05 & 1.97 & 1.42 & 10.94 \\
        w/ OC & 43.71 & 15.36 & 24.21 & \textbf{96.30} & 83.82 & \textbf{2.76} & \textbf{3.94} & \textbf{19.63} \\
        w/ Dual & \textbf{44.96} & \textbf{15.92} & \textbf{24.97} & 93.24 & 86.12 & 2.61 & 2.85 & 15.97 \\
        w/ OC + Dual & 42.77 & 14.89 & 23.74 & 95.71 & \textbf{86.59} & 2.49 & 3.29 & 17.68 \\
        \midrule
        ChatGPT-3.5-turbo & 46.36 & 16.47 & 26.13 & 98.64 & 84.34 & 2.37 & 2.01 & 16.93 \\
        w/ OC & 46.74 & 16.70 & 26.58 & 99.17 & 85.70 & \textbf{2.86} & 4.64 & \textbf{20.11} \\
        w/ Dual & 47.42 & 17.12 & 27.30 & 99.25 & \textbf{86.34} & 2.71 & 3.85 & 19.07 \\
        w/ OC + Dual & \textbf{47.61} & \textbf{17.15} & \textbf{27.48} & \textbf{99.31} & 86.22 & 2.85 & \textbf{4.76} & 19.40 \\
        \midrule
        Ground Truth & - & - & - & 95.69 & - & 2.89 & 6.73 & 22.63 \\
        \bottomrule
    \end{tabular}}
    \caption{Precise outline-conditioned generation performance of the base models and after applying our proposed \textbf{explicit outline utilization control (OC)}, \textbf{unified dual-task learning (Dual)}, and their combination \textbf{(OC+Dual)}. The best results for each metric in each group are in \textbf{bold}.}
    \label{tab:overall}
\end{table*}

\subsubsection{Main Results}

Table~\ref{tab:overall} shows the main results of precise outline-conditioned generation on the two datasets CDM and WPOG under both fine-tuning and zero-shot inference settings. We report results based on the automatic evaluation metrics described in Section~\ref{sec:metric}. 
For a fine-tuning baseline, we use the encoder-decoder BART-base\footnote{https://huggingface.co/facebook/bart-base}; for zero-shot inference baseline, we use ChatGPT-3.5-turbo\footnote{https://chat.openai.com}. 
For each baseline, we apply our proposed methods, \textbf{explicit outline utilization control (OC)}, \textbf{unified dual-task learning (Dual)}, and their combination \textbf{(OC+Dual)}.


In the fine-tuning setting, we find \textbf{both OC and Dual improve the baseline across all metrics on both datasets}, confirming the effectiveness of our methods. Notably, OC achieves a larger gain than Dual on CDM, especially on the diversity metric D-4, because the CDM outline reveals less information about the order and correspondence between points and news content. Since OC helps convey this information, it achieves larger gains than Dual on CDM. In contrast, OC has a smaller impact on WPOG than on CDM, because WPOG outlines are more explicit about the key plots and story elements. There is a clearer correspondence between paragraphs in the story and key points in the outline in WPOG, which likely reduces the benefit of OC. 
Notably, OC achieves substantial improvements on DV, PD, and CD on both CDM and WPOG, demonstrating the remarkable effectiveness of OC in facilitating more even outline utilization.
 
Combining OC and Dual (OC+Dual) improves the baseline, but it does not achieve the best results on most of the metrics compared to each individual approach. We think this is probably because of the conflict between the two methods when they are combined. To ensure that the dual loss from the summarization task can pass the relevant parts to the separate section in the generation under OC, we applied the same partitioning method to the summarization task, transforming it from a full-passage summarization to a partial point-wise summarization task, which may hurt the completeness of the summarization task.

In the zero-shot inference setting, \textbf{both OC and Dual substantially improve the baseline performance on both datasets, with Dual outperforming OC}. Compared with the baseline used in fine-tuning, LLMs have stronger power in contextual understanding. We believe this makes LLMs benefit more from restructuring and revising, rather than teaching them to generate segment by segment directly. The combination of OC+Dual further improves R-1 and PD metrics on both datasets and achieves comparable results on the other metrics, compared to using OC and Dual individually. 

\subsubsection{Analysis}
\paragraph{\textbf{Impact of model size and architecture for fine-tuning scenario.}} We compare the performance of encoder-decoder models BART-base (139M parameters) and BART-large (406M parameters) with decoder-only models GPT-2-base (124M parameters) and GPT-2-large (774M parameters) for fine-tuning under outline-conditioned generation, as shown in Table \ref{tab:ft_size}. Both OC and Dual improve all the baseline models regardless of their model size and architecture. 
When comparing OC and Dual, we observe that BART-base and GPT-2-base with OC achieve better results than Dual on all metrics except CD. This observation highlights that both encoder-decoder and decoder-only models exhibit an imbalanced outline utilization issue, and it underscores the effectiveness of OC in mitigating this concern.


\begin{table}[htb]
\renewcommand{\arraystretch}{1}
\scalebox{0.7}{
\begin{tabular}{lccccccc}
\toprule
Method      & R-1            & R-2            & R-L            & DV            & PD            & CD             \\
\midrule      
BART-base        & 38.79          & 10.86          & 16.26          & 2.21          & 1.09          & 14.74          \\
w/ OC       & \textbf{40.92}          & \textbf{11.82}          & \textbf{17.48}          & \textbf{2.94} & \textbf{4.06} & \textbf{21.62}          \\
w/ Dual     &39.66           & 11.21          & 16.67          & 2.43          & 2.37  &15.20 \\
\midrule
BART-large  & 40.26          & 10.75          & 17.05          & \textbf{2.51}          & 2.78          & 16.28          \\
w/ OC       & \textbf{41.39} & \textbf{12.06} & \textbf{18.43} & 2.46          & \textbf{3.02}          & \textbf{23.81} \\
w/ Dual     &40.41           &11.69           &17.74           & 2.36          &2.96  &19.46 \\
\midrule
GPT-2-base      & 34.63          & 8.29           & 13.48          & 1.92          & 1.21          & 9.76           \\
w/ OC       & \textbf{36.92}          & \textbf{10.22}          & \textbf{15.53}          & \textbf{2.64}          & \textbf{3.39}          & 15.94          \\
w/ Dual    & 36.48           & 10.19          & 15.19          & 2.30          & 2.90       &\textbf{16.24}  \\
\midrule
GPT-2-large & 37.95          & 9.30           & 15.86          & 2.35          & 1.74          & 12.57          \\
w/ OC       & 38.20          & 10.25          & 16.14          & \textbf{2.76}          & \textbf{3.42}          & 14.09            \\
w/ Dual     & \textbf{38.87}          &\textbf{10.69}           &\textbf{16.49}           &2.49           &2.19        &\textbf{15.78}  \\ 
\bottomrule
\end{tabular}}
\caption{Impact of model sizes and architectures of fine-tuning LMs for precise outline-conditioned generation on CDM. The best results for each metric in each group are in \textbf{bold}.}
\label{tab:ft_size}
\end{table}

\begin{table}[htb]
\renewcommand{\arraystretch}{1}
\scalebox{0.72}{
\begin{tabular}{lccccccc}
\toprule
Method     & R-1            & R-2            & R-L            & DV            & PD            & CD             \\
\midrule
ChatGPT    & 41.17          & 12.17          & 17.62          & 2.63          & 3.44          & 18.96          \\
w/ OC      & 41.69          & 12.78          & 18.67          & \textbf{2.75} & \textbf{4.46} & \textbf{23.88} \\
w/ Dual    & \textbf{42.46} & \textbf{13.10} & \textbf{19.06} & 2.49          & 3.62          & 18.54          \\
\midrule
Vicuna-13B & 39.43          & 11.28          & 17.92          & 2.14          & 2.90          & 16.83          \\
w/ OC      & \textbf{41.26}          & 11.74          & \textbf{18.20}          & 2.36          & 3.17          & \textbf{19.07}          \\
w/ Dual    & 40.90          & \textbf{12.42}          & 18.07          & \textbf{2.42}          & \textbf{3.29}          & 18.29          \\
\bottomrule
\end{tabular}}
\caption{Impact of using different LLMs under zero-shot inference setting for precise outline-conditioned generation on CDM. The best results for each metric in each group are in \textbf{bold}.}
\label{tab:zs_size}
\end{table}


\paragraph{\textbf{Impact of different LLMs for zero-shot inference scenario.}}We compare the performance of Vicuna-13B and ChatGPT for zero-shot inference under precise outline-conditional generation, as shown in Table \ref{tab:zs_size}. The results show that \textbf{both OC and Dual improve} the performance of both baseline models on all metrics and datasets. This indicates that \textbf{our methods are effective} for different kinds of LLMs under zero-shot inference scenarios. 
Unlike ChatGPT, Vicuna-13B benefits more from OC on R-1 and R-L, which suggests that OC works better on smaller models.\footnote{The impact of in-context learning is shown in Appendix.~\ref{sec:analysis_appendix}.}

\begin{figure}[ht]
    \setlength{\abovecaptionskip}{0pt} 
    \setlength{\belowcaptionskip}{0pt} 
    \centering  
    \subfigure[BART]{
        \includegraphics[width=0.35\textwidth]{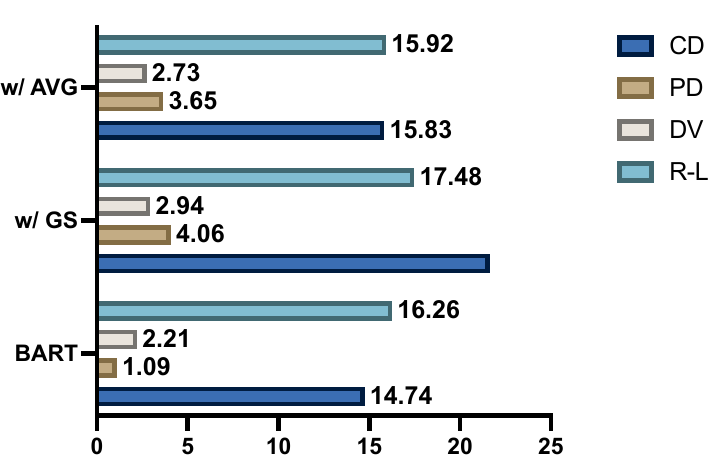}}
    \subfigure[GPT-2]{
        \includegraphics[width=0.35\textwidth]{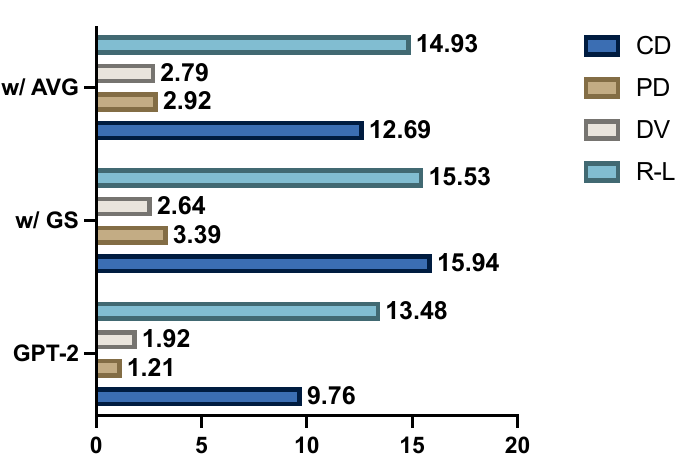}}
    \caption{The impact of different partition methods on the performance of \textbf{explicit outline utilization control (OC)} on CDM.}
    \label{fig:par}
\end{figure}

\paragraph{Impact of partition methods for outline control in fine-tuning scenario.} We propose two ways (average partition and greedy search) to make partitions for stories, as mentioned in section \ref{sec:method}. We compare the performance of these two ways on BART-base and GPT-2-base for fine-tuning scenarios. The results are shown in Figure \ref{fig:par}. The performance of greedy search \textbf{significantly outperforms} the average partition, which indicates that the suitable partition contributes to the generation performance by establishing the mapping between outlines and parts of stories. However, the different partitioning methods show consistent improvements in outline-related metrics (i.e., DV, PD, and CD) compared to the baseline model. This demonstrates the effectiveness of outline control in mitigating the problem of imbalanced utilization of the outline information.

\subsection{Human Evaluation}
Besides the validation of the WPOG dataset, we conduct another two human evaluation experiments to answer the following research questions: 

\noindent\textbf{RQ1:} How is the text generation quality of the models evaluated in this task in two different scenarios? 

\noindent\textbf{RQ2:} Can the new metric proposed in this paper correctly reflect the usage of outlines?

Following the setting of previous works ~\citep{Yang2023DOCIL, zhou2023recurrentgpt}, we sample 20 outlines and compare corresponding stories generated by the base and OC+Dual variants in both fine-tuning and zero-shot inference. We ask ten evaluators to rate the quality of stories in different categories: fluency, coherence and flow, completeness,	relevance, beginning, closure, and outline ordering. The details about criteria and questions are shown in  Figure~\ref{fig:question_result_eval} (Appendix~\ref{sec:questionnaire}).

\begin{figure}[ht]
    \centering
    \includegraphics[width=0.49\textwidth]{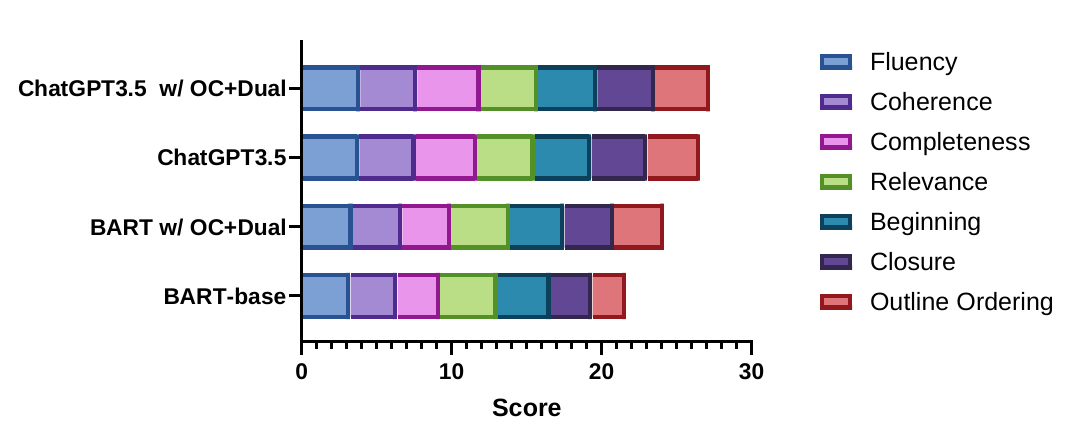}
    \caption{The human evaluation on overall score and detailed performance of the four methods.}
    \label{fig:human_eval}
\end{figure}

In Figure~\ref{fig:human_eval}, we show the overall score and detailed performance of the four methods. It is observed that variants employing OC+Dual strategies have shown improvements compared to the base model, which is consistent with the trend reflected by automatic evaluation results. It is notable that, for fine-tuning inference, the variant gets strong improvement on two structure metrics, i.e., closure and outline ordering. It demonstrates that the strategies we proposed can alleviate the issue that the outline information is overused at the beginning. In terms of faithfulness, the variants achieve 0.5 and 0.18 points higher than the base model in the combination of completeness and relevance over fine-tuning and zero-shot inference, respectively.

\begin{figure}[ht]
    \centering
    \setlength{\belowcaptionskip}{-0.5cm} 
    \includegraphics[width=0.4\textwidth]{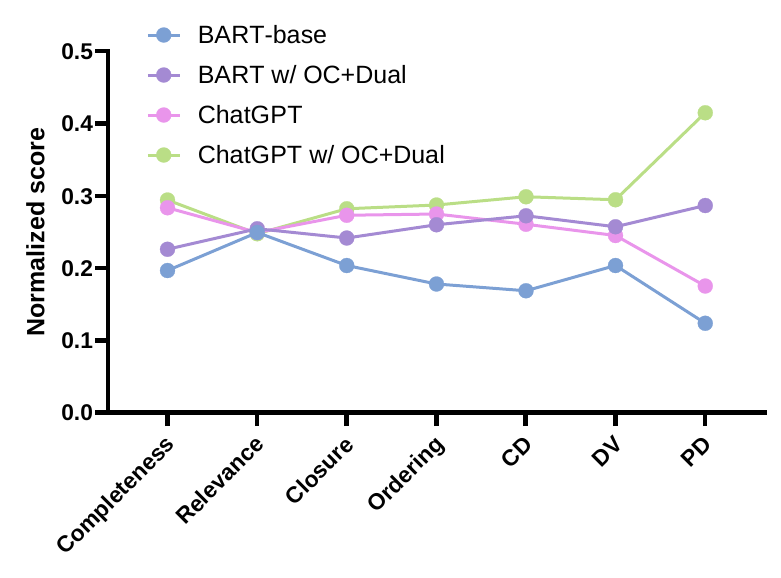}
    \caption{Comparison among human evaluation and automatic metrics.}
    \label{fig:human_eval_metric}
\end{figure}

For RQ2, we compare the results of our proposed outline usage automatic metrics: Peak-value Distance (PD), Distribution Variation (DV), and Consistency Degree (CD), and four outline-relevant human evaluation metrics: completeness, relevance, closure, and outline ordering as shown in Figure \ref{fig:human_eval_metric}. In overall comparison,  they show similar trends in distinguishing the base model and their corresponding variants. The result on extrinsic metric CD is close to outline ordering, which demonstrates the way to reconstruct the outline could reflect the structure character of generated text. When it comes to intrinsic metrics, the variation in the distribution of outline similarity demonstrates greater robustness in comparison to the distance between sentences that are most similar to the outline. This robustness aligns better with human evaluation metrics.

\section{Conclusion}
In this paper, we introduce a novel text generation task called precise outline-conditioned generation, which generates stories based on specific, sentence-level outlines. We construct two new benchmark datasets, WPOG and CDM, for this task. We provide strong baselines based on fine-tuning models such as BART and GPT-2, and evaluating zero-shot performance of models such as ChatGPT and Vicuna. We identify an issue related to the imbalanced utilization of the outline information in the precise outline-conditioned generation, which affects both fine-tuning and zero-shot models. To address this issue, we have proposed a unified explicit outline utilization control approach and a novel framework that leverages the task duality between summarization and generation. We have shown that our proposed approaches effectively alleviate the issue of imbalanced outline utilization and enhance the quality of precise outline-conditioned text generation for both fine-tuning and zero-shot settings. We hope that our work will inspire more research on this challenging and interesting task.


\section{Limitations}
As discussed in this paper, the proposed task of precise outline-conditioned generation could be regarded as a key subtask in conditional or open-ended long-text generation. However, we only evaluate our methods on texts with a length limitation of one thousand tokens (approximately 750 words). We do not assess the performance of our strategies on longer texts (thousands of words or even longer) since the quality of long text generation heavily depends on human evaluation, which is time-consuming and expensive.
\section{Ethical Considerations}

Our work focuses on advancing outline-conditioned text generation technology. We acknowledge that similar technology can be utilized to generate deceptive or manipulative content such as fake news. Our research provides unified strategies that are utilized by base language models, compatible with their detection and restriction of harmful information and false information. Additionally, the result demonstrates our method could improve the faithfulness of generated text to the given outline, making it easy for further detection and control over the input outline rather than whole articles. The zero-shot inference prompts are currently tailored exclusively for the English language, and extending them to other languages would necessitate further adjustments.  The effectiveness of our methods could potentially be compromised in languages with fewer resources, as we rely extensively on pretrained language models that may exhibit diminished performance in such linguistic contexts. 

\bibliography{custom}
\bibliographystyle{acl_natbib}

\appendix

\section{Prompt Design}
Here we list the prompt we used under the zero-shot scenario for reproducing our experiment results and the one we used in evaluating consistent degree (CD). Here we take the prompt on CDM as example, while WPOG only replaces the ``news report'' by ``story''. It is noticeable that both ChatGPT and Vicuna are sharing these prompts.
\label{sec:prompt}
\subsection{Prompt - Baseline}
\label{subsec:allin}
\noindent {\fontfamily{bch}\selectfont
\textbf{User}: Please generate a [XXX] words news report with given first sentence and outlines. The first sentence is: [XXX]. And the outline of this news report is [XXX]. 

\vspace{2mm}
\noindent \textcolor{blue}{\textbf{Agent}: [XXX].}
}

\subsection{Prompt - Explicit Outline Utilization Control (OC)}
\label{sec:separate}

\noindent {\fontfamily{bch}\selectfont
\textbf{User}: Your final goal is to generate a [XXX] words news report based on given first sentence and outlines. The first sentence is: [XXX] And the outline of this news report is: [XXX]. Firstly, you could generate a part of the news report corresponding to following plot [XXX].

\vspace{2mm}
\noindent \textcolor{blue}{\textbf{Agent}: [XXX].}
\vspace{2mm}

\noindent \textbf{User}: Continue to generate a part of the news report followed by your previous output while the plot is corresponding to the following plot [XXX].

\vspace{2mm}
\noindent \textbf{\textcolor{blue}{\textbf{Agent}} \& User}: ... repeat m-1 times 
\vspace{2mm}

\noindent \textbf{User}: Now connect all the paragraphs you've written and polish them to a [XXX] tokens essay to achieve the final goal to generate an around [XXX] tokens news report with given first sentence and outlines. First sentence: [XXX]  Outline: [XXX].
\vspace{2mm}

\vspace{2mm}
\noindent \textcolor{blue}{\textbf{Agent}: [XXX].}
\vspace{2mm}

\subsection{Prompt - Unified Dual-task Learning (Dual)}
\label{subsec:dual}
{\fontfamily{bch}\selectfont
\noindent \textbf{User}: Please generate a [XXX] words news report with given first sentence and outlines. The first sentence is: [XXX]. And the outline of this news report is [XXX]. 

\vspace{2mm}
\noindent \textcolor{blue}{\textbf{Agent}: [XXX].}
\vspace{2mm}

\noindent \textbf{User}: Please summarize it into [XXX] one-sentence points.

\vspace{2mm}
\noindent \textcolor{blue}{\textbf{Agent}: [XXX].}
\vspace{2mm}

\noindent \textbf{User}: Please compare with the following true outline and rethinking how to improve the quality of outline-conditioned news report generation: [XXX].

\vspace{2mm}
\noindent \textcolor{blue}{\textbf{Agent}: [XXX].}
\vspace{2mm}

\noindent \textbf{User}: Based on the knowledge you just learned, regenerate a [XXX] words news report with given first sentence and outlines. The first sentence is: [XXX]. And the outline of this news report is [XXX].}


\subsection{Prompt - Combo: OC + Dual }
\label{subsec:combo}
{\fontfamily{bch}\selectfont
*** Repeat all steps in OC mode ***

\noindent \textbf{User}: Please generate a [xxx] tokens story with the given first sentence and outlines. The first sentence is: [XXX]. And the outline of this story is [XXX]

\vspace{2mm}
\noindent \textcolor{blue}{\textbf{Agent}: [XXX].}
\vspace{2mm}

\noindent \textbf{User}: Please summarize it into an outline that has [XXX] one-sentence points.

\vspace{2mm}
\noindent \textcolor{blue}{\textbf{Agent}: [XXX].}
\vspace{2mm}

\noindent \textbf{User}: Please compare with the following true outline and rethinking how to improve the quality of outline-conditioned news report generation. True outline: [XXX].

\vspace{2mm}
\noindent \textcolor{blue}{\textbf{Agent}: [XXX].}
\vspace{2mm}

\noindent \textbf{User}: Based on the knowledge you just learned, regenerate a [XXX] words news report with given first sentence and outlines. The first sentence is: [XXX]. And the outline of this news is [XXX].}

\noindent \textbf{User}: Refine your output to a [XXX] words news report. 
}

\subsection{Prompt - Consistency Degree (CD)}
\label{subsec:cd}
{\fontfamily{bch}\selectfont
\textbf{User}: Please generate an outline for the given news report. The news report: [XXX]
\vspace{2mm}

\noindent \textcolor{blue}{\textbf{Agent}: [XXX].}
\vspace{2mm}
}
\section{Experimental Setup}
\label{sec:aes}
\paragraph{Data Preprocessing} For both two datasets, we use the first sentence as input. While CNN/DailyMail usually includes some meaningless openings for news reports (e.g., reporter's name), we made an additional filter for this issue, which only accepts openings with more than 7 words. Moreover, we aim to generate long text involving abundant semantic and structural information. Here we filter out all cases with less than 64 words. We cut off 40 sentences to test model performance. 

\paragraph{Implementation Details} Our code implementations are mainly based on Pytorch 2.0 and the Huggingface library. All experiments including model finetuning (e.g., BART, GPT-2) and zero-shot inference (e.g., Vicuna) are running on 8 NVIDIA V100 32G GPUs. The total computation cost is about 1500 GPU hours. For fine-tuning BART and GPT-2, we set the batch size as 4, and training epochs as 3 on CDM, and 20 on WPOG since they have different scales of training set. The hyperparameters $\alpha,\beta$ in outline control are set to 0.8, and 0.05, which are obtained by grid search from $\{0.5,0.6,0.7,0.8,0.9\}$ and $\{0.005,0.01,0.05,0.1\}$. We report the mean value for all results after running the experiments 3 times. It is noticeable that the structure of GPT-2 is different from BART, we concatenate both the opening and outline together through prompts as input. For zero-shot scenarios, we choose two state-of-the-art large language models. For the open-source model, we choose Vicuna-v1.5-13B\footnote{https://huggingface.co/lmsys/vicuna-13b-v1.5}, which is derived from Llama-2~\citep{touvron2023llama}. It achieved \#2 in Arena Elo rating among all 13B parameter LLMs in Massive Multitask Language Understanding (MMLU). For the close-source model, we use ChatGPT by OpenAI's 3.5-turbo API. For dataset creation, we use GPT-4\footnote{https://openai.com/gpt-4} API to generate the outlines. According to the official price and our usage, we spend around \$700 on ChatGPT API costs.

For human evaluation, we recruited 10 graduate students from the university as volunteers. They are from the United States and China. We provide them with \$5 for dataset verification and \$15 for outline-conditioned generation evaluation.

\section{More Result Analysis}
\label{sec:analysis_appendix}
\paragraph{\textbf{Performance over in-context learning }}
To further evaluate the effectiveness of the proposed strategies in the zero-shot inference scenario, we compare them with the application of in-context learning to the base method by sampling another case and adding it to the prompt.

\begin{table}[ht]
\renewcommand{\arraystretch}{1}
\scalebox{0.78}{
\begin{tabular}{lccccc}
\toprule  
\textbf{Method} & \textbf{R-1}   & \textbf{R-2}   & \textbf{R-L}   & \textbf{D-4}   & \textbf{BS} \\
\midrule
ChatGPT & 46.36 & 16.47 & 26.13 & 98.64 & 84.34 \\
w/ in-context         & 46.53 & 16.39 & 26.22 & 98.31 & 84.40 \\
w/ OC                        & 46.74 & 16.70 & 26.58 & 99.17 & 85.70 \\
w/ Dual         & \textbf{47.42} & \textbf{17.12} & \textbf{27.30} & \textbf{99.25} & \textbf{86.34}  \\   
\bottomrule
\end{tabular}}
\caption{Impact of in-context learning on ChatGPT over WPOG dataset}
\label{tab:in}
\end{table}

As shown in Table~\ref{tab:in}, it is evident that both OC and Dual consistently outperform in-text learning. Even when comparing in-text learning to the base model, it becomes apparent that in-text learning does not yield a substantial improvement (lower performance on Rouge-2 and Distinct-4). We hypothesize that this lack of improvement may be attributed to the random selection of samples, which might introduce stylistic variations differing from the original cases, thereby causing confusion for the model. This observation also underscores the effectiveness of our proposed strategies.

\begin{figure*}[ht]
  \centering
  \subfigure[]{
    \includegraphics[width=0.48\textwidth]{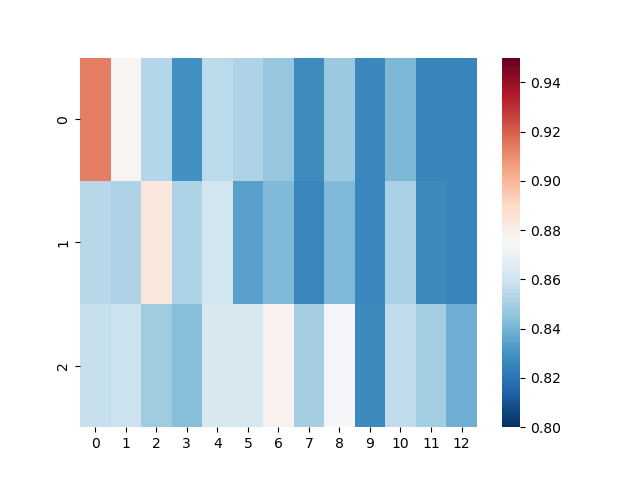}
    \label{fig:gt_dis}
  }
  \hfill
  \subfigure[]{
    \includegraphics[width=0.48\textwidth]{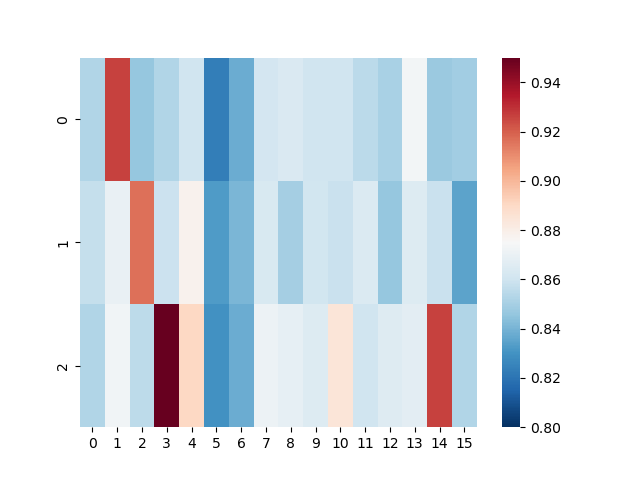}
    \label{fig:allin_dis}
  }
  
  \subfigure[]{
    \includegraphics[width=0.48\textwidth]{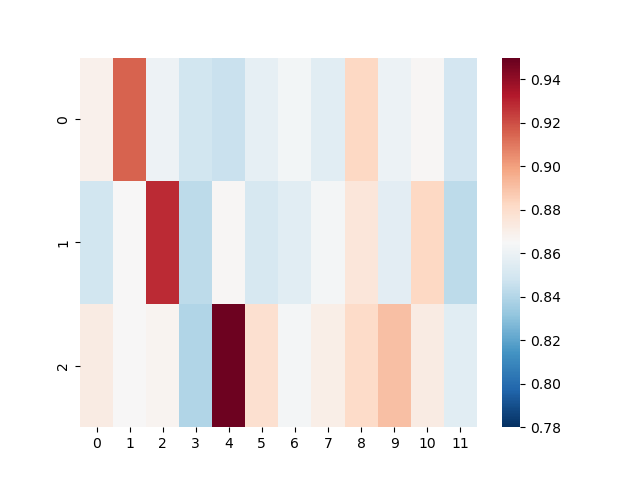}
    \label{fig:llm_dis}
  }
  \hfill
  \subfigure[]{
    \includegraphics[width=0.48\textwidth]{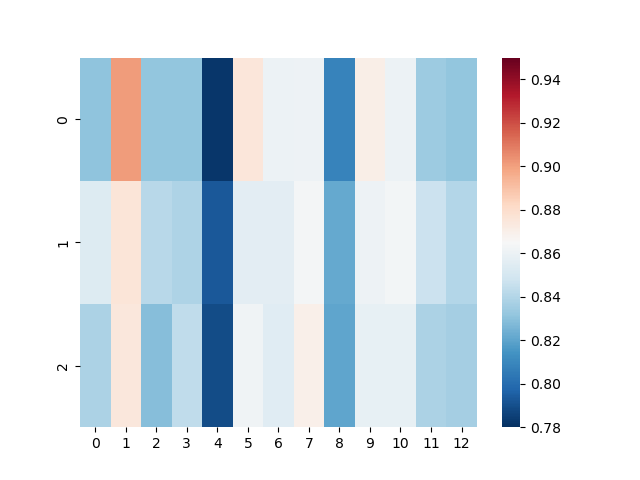}
    \label{fig:seperate_dis}
  }

  \caption{The case study of outline similarity within the entire story in CDM. (a) Ground truth text, (b) Generated text from BART-base without outline control, (c) Generated text from ChatGPT (gpt-3.5-turbo) without outline control, (d) Generated text from BART-base with outline control. The Y-axis represents the outline bullet points, while the X-axis represents the sentences in the generated text. The values indicate the corresponding similarity between the outline bullet point and the generated sentences. }
  \label{fig:dis}
\end{figure*}

\paragraph{\textbf{Case study about imbalanced outline information usage}} The similarity measurement between the outline and the entirety of the generated text provides a clear illustration of this issue, as the case study depicted in Figure \ref{fig:dis}. In the text generated by BART (Figure \ref{fig:allin_dis}), the utilization of information is primarily focused at the beginning, contrasting with the distribution observed in the ground truth (Figure \ref{fig:gt_dis}). Furthermore, a parallel trend is observed in large language models (LLMs), such as ChatGPT (Figure \ref{fig:llm_dis}), signifying that the imbalanced utilization of outlines is a pervasive issue in text generation. The experimental results, depicted in Figure \ref{fig:seperate_dis}, validate the efficacy of this approach. 

\section{Dataset Quality Validation}
\label{sec:data_valid}
We evaluate the quality of AI-annotated outlines by accuracy, completeness, and clarity. The definition and criteria are shown in Figure~\ref{fig:question_data_eval} (Appendix~\ref{sec:questionnaire}). 
Specifically, we randomly select 20 pairs of <story, outline> from WPOG and another 20 pairs from CDM. Each evaluator is asked to rate from 1 to 5 for a sample of 5 stories from each of the two datasets. To avoid bias, we ensure that each sample is rated by two different evaluators. 

\begin{figure}[ht]
    \centering
    \includegraphics[width=0.49\textwidth]{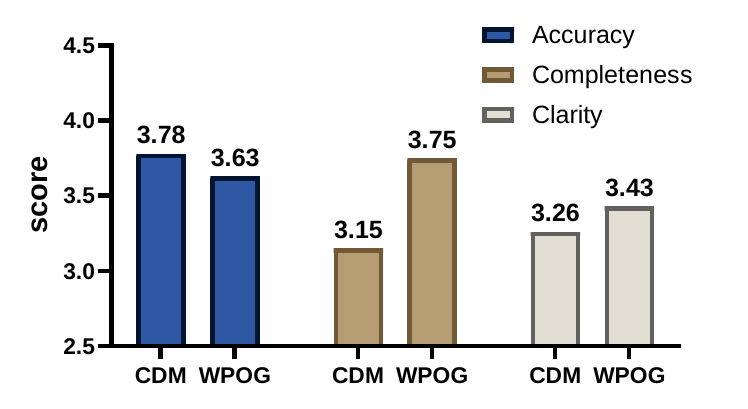}
    \caption{The human evaluation on dataset quality validation.}
    \label{fig:data_valid}
\end{figure}

As Figure~\ref{fig:data_valid} illustrates, the WPOG's outline achieves higher scores in both completeness and clarity. while it slightly lags behind CDM in terms of accuracy. We believe this is due to the fact that the model adds some details to maintain the structural integrity of the outlines, thus sacrificing some accuracy. Taking these three metrics into account, the quality of the WPOG dataset is compelling.

\section{Examples of Precise Outline-conditioned Generation}
\label{sec:example}
Table~\ref{tab:example} shows an example of how the baseline models of BART and ChatGPT generate stories based on prompts and outlines.
\label{sec:example_obg}
\begin{table*}
\centering
\scalebox{0.72}{
\begin{tabular}[]{p{0.5in}p{7.5in}}
\toprule
\multicolumn{2}{p{5in}}{
\textbf{Examples of precise outline-conditioned generation}:}\\
\midrule
\textbf{Outline} & \colorbox{yellow}{Tourists and locals queue for several hours to get their hands on Jenny's butter cookies .}\newline
\colorbox{pink}{People are even hired to stand in line to buy the cookies, which are later sold at an up-to-70\% mark-up }\newline
\colorbox{green}{Food frenzies have also taken place in other parts of the world .} \\
\midrule
\textbf{Prompt} & There's a booming black market in Hong Kong, but it's not for fake Apple Watches, or the iPhone. \\
\midrule
\textbf{Output (BART)} & There's a booming black market in Hong Kong, but it's not for fake Apple Watches or iPhones. \colorbox{yellow}{Tourists and locals queue for hours to buy Jenny's butter cookies.}\colorbox{pink}{People are even hired to stand in line and later} \colorbox{pink}{sell the cookies at a hefty markup, sometimes as much as 70\%.}  \colorbox{green}{These food frenzies aren't limited to Hong Kong;}  \colorbox{green}{they're happening worldwide.}  From limited-edition Kit Kats in Japan to bagel crazes in New York, food tourism is influencing travel.  This trend raises questions about what drives such desires for specific foods and what it means for the culinary and tourism industries.  Jenny's butter cookies remain a testament to the power of taste and the lengths people go for a delicious treat.\newline\newline
\colorbox{yellow}{Jenny's butter cookies, with their irresistible allure, serve as a sweet testament to the lengths people will go to} \colorbox{yellow}{  for a bite of something truly special.} These worldwide culinary obsessions beg the question: What fuels these fervent desires for specific foods? The answer may lie in the power of taste and the universal appeal of exceptional culinary experiences.\colorbox{pink}{Once secured, these cookies are later resold at an astonishing markup,} \colorbox{pink}{sometimes reaching up to 70\% above the original price.} It's a business strategy that has tapped into the undeniable allure of Jenny's cookies and the willingness of customers to pay a premium for them. Queueing for hours has become a common sight as both locals and tourists are eager to savor these delectable treats. But this craze doesn't stop at the cookie shop's door.
\\

\midrule
\textbf{Output (ChatGPT):} &There's a booming black market in Hong Kong, but it's not for fake Apple Watches or iPhones.The latest craze sweeping the vibrant city isn't about tech gadgets;   it's about butter cookies. \colorbox{yellow}{Jenny's butter cookies, to be precise, have become a culinary sensation, with both locals and tourists eagerly } \colorbox{yellow}{joining long queues for a taste of these delectable treats.} The scene is surreal. \colorbox{pink}{ People stand in line for several }  \colorbox{pink}{hours , determined to get their hands on Jenny's butter cookies.} The allure of these sweets is irresistible, with their reputation for unmatched flavor and melt-in-your-mouth goodness. Tourists visiting Hong Kong are often captivated by the buzz surrounding these cookies, making them an essential item on their travel itineraries. \colorbox{green}{But Hong Kong isn't the only place where food frenzies have taken hold.} \newline

However, what's intriguing is how this intense demand has given birth to a clandestine world of commerce. To meet the soaring demand for these cookies, some individuals are hired to stand in line and purchase the coveted treats, which are later sold at a staggering markup. \colorbox{pink}{Reports suggest that these cookies, originally priced } \colorbox{pink}{reasonably, are now reselling for up to a 70\% increase in value.} It's a clear reflection of the extent to which people are willing to go for a taste of this newfound delicacy.\newline

Across the globe, similar phenomena have unfolded. From Taiwan's famous pineapple cakes to Parisian macarons, food enthusiasts are demonstrating an insatiable appetite for unique and sought-after culinary experiences, prompting culinary tourism to flourish.\newline

As the butter cookie craze continues to captivate hearts and taste buds, it raises questions about the nature of trends in the modern world. What drives the fervent desire for specific foods, and what do these food frenzies mean for the culinary industry and global tourism? In the midst of it all, Jenny's butter cookies remain a sweet testament to the power of taste, the lengths people will go for a delicious bite, and the intriguing world of culinary commerce.\\
\bottomrule
\end{tabular}
}
\caption{Examples of precise outline-conditioned generation over CDM. The highlights reflect the usage of corresponding outline information}
\label{tab:example}
\end{table*}


\section{Questionnaire for Human Evaluation}
\label{sec:questionnaire}

\begin{figure}[ht]
  \centering
  \subfigure[]{
    \includegraphics[width=0.45\textwidth]{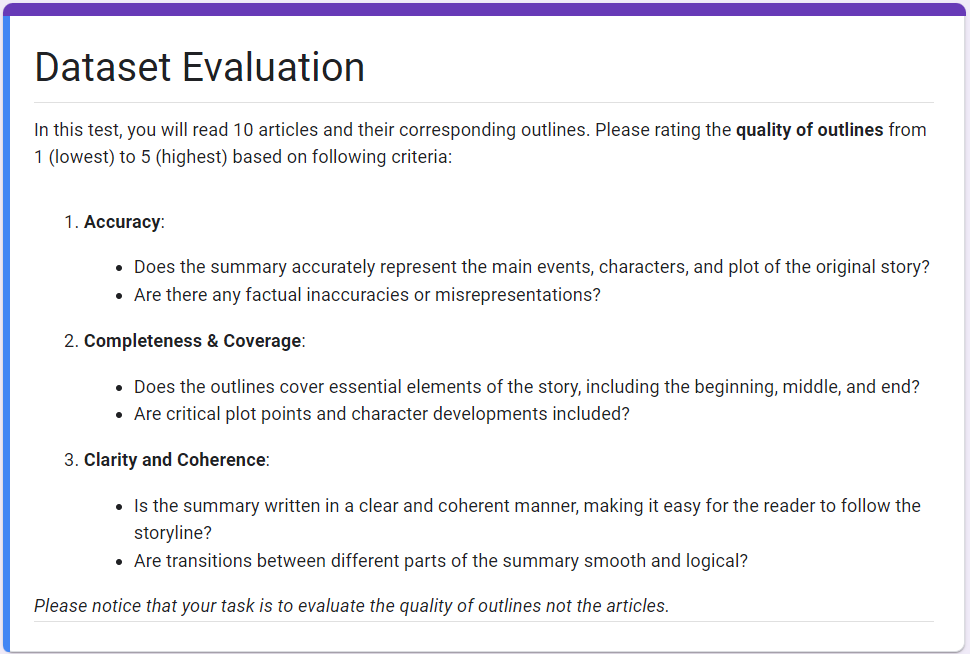}
  }
  \hfill
  \subfigure[]{
    \includegraphics[width=0.45\textwidth]{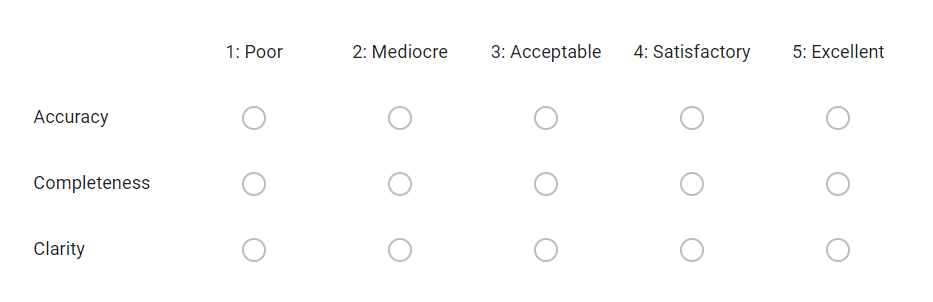}
  }

  \caption{An example about explanation (a) and questions (b) in the questionnaire of dataset validation from human evaluation.}
  \label{fig:question_data_eval}
\end{figure}

\begin{figure}[ht]
  \centering
  \subfigure[]{
    \includegraphics[width=0.45\textwidth]{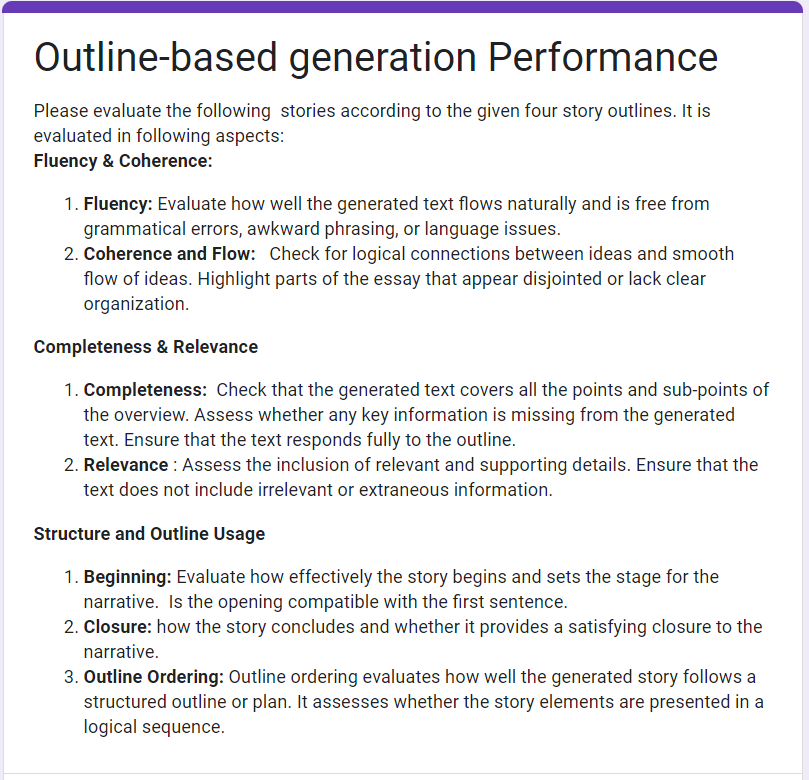}
  }
  \hfill
  \subfigure[]{
    \includegraphics[width=0.45\textwidth]{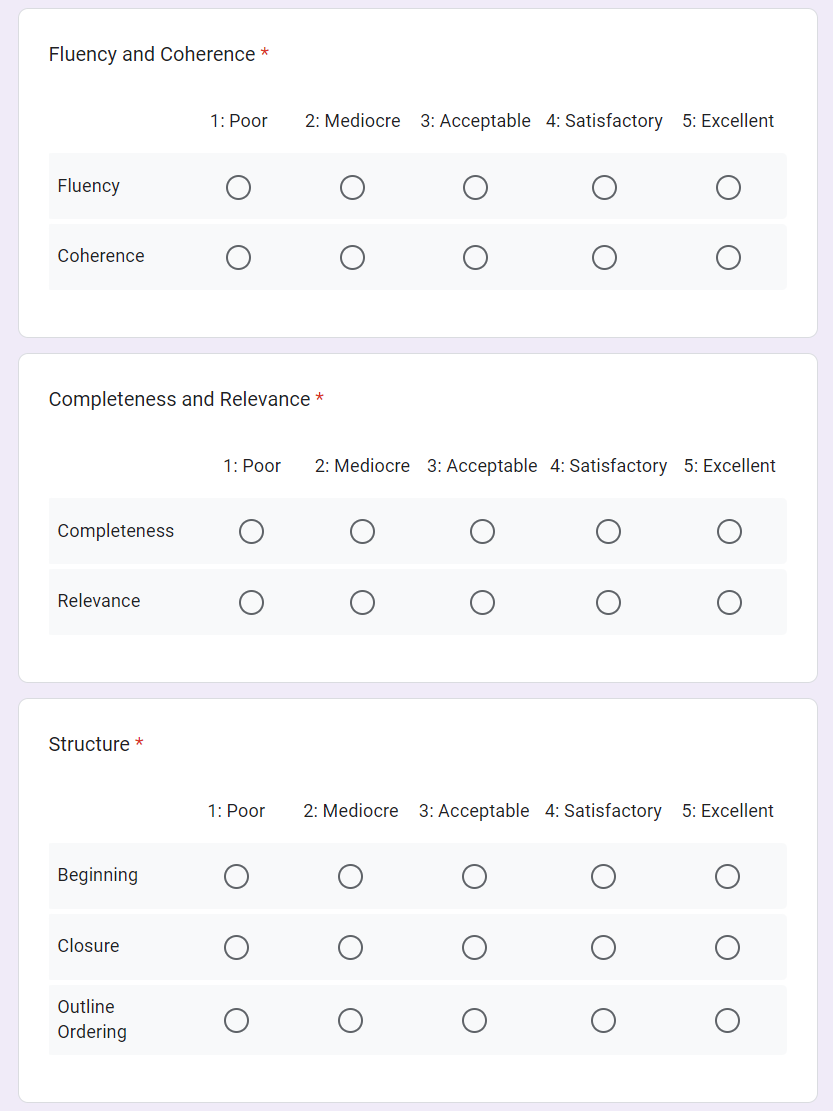}
  }

  \caption{An example about explanation (a) and questions (b) in the questionnaire of the quality of generated stories from human evaluation.}
  \label{fig:question_result_eval}
\end{figure}

The questionnaire for dataset validation and quality of generated stories consists of explanations and questions, as shown in Figure \ref{fig:question_data_eval} and Figure \ref{fig:question_result_eval}, respectively. 

\end{document}